%% file: clear2023.tex
\documentclass[final,12pt]{clear2023} % Include author names

\title[]{Towards a Solution to Bongard Problems: A Causal Approach}
\usepackage{times}
\usepackage{wrapfig}
\usepackage[most]{tcolorbox}

\author{\textbf{Salahedine Youssef}\textsuperscript{\rm 1}
\quad
\textbf{Matej Zečević}\textsuperscript{\rm 1}
\quad \textbf{Devendra Singh Dhami}\textsuperscript{\rm 1, 3} \quad \textbf{Kristian Kersting}\textsuperscript{\rm 1-4}\\
{\normalfont \textsuperscript{\rm 1}Computer Science Department, TU Darmstadt, \textsuperscript{\rm 2}Centre for Cognitive Science, TU Darmstadt, \textsuperscript{\rm 3}Hessian Center for AI (hessian.AI),
\textsuperscript{\rm 4}DFKI}
}

\begin{document}

\maketitle

\begin{abstract}%
  Even though AI has advanced rapidly in recent years displaying success in solving highly complex problems, the class of Bongard Problems (BPs) yet remain largely unsolved by modern ML techniques. In this paper, we propose a new approach in an attempt to not only solve BPs but also extract meaning out of learned representations. This includes the reformulation of the classical BP into a reinforcement learning (RL) setting which will allow the model to gain access to counterfactuals to guide its decisions but also explain its decisions. Since learning meaningful representations in BPs is an essential sub-problem, we further make use of contrastive learning for the extraction of low level features from pixel data. Several experiments have been conducted for analyzing the general BP-RL setup, feature extraction methods and using the best combination for the feature space analysis and its interpretation. \par
\end{abstract}


\input{master_thesis/1_Introduction}
\input{master_thesis/2_Related_Work}
\input{master_thesis/3_Methodology}

\input{master_thesis/5_Analysis_Part_II}

\input{master_thesis/6_Conclusion}

\section*{Acknowledgements}
The authors acknowledge the support of the German Science Foundation (DFG) project “Causality, Argumentation, and Machine Learning” (CAML2, KE 1686/3-2) of the SPP 1999 “Robust Argumentation Machines” (RATIO). This work was supported by the ICT-48 Network of AI Research Excellence Center “TAILOR” (EU Horizon 2020, GA No 952215), the Nexplore Collaboration Lab “AI in Construction” (AICO) and by the Federal Ministry of Education and Research (BMBF; project “PlexPlain”, FKZ 01IS19081). It benefited from the Hessian research priority programme LOEWE within the project WhiteBox \& the HMWK cluster project “The Third Wave of AI” (3AI).

\bibliography{uai2022crl.bib}

\appendix

\input{master_thesis/appendix}

\end{document}

%% file: master_thesis/1_Introduction.tex
\section{Motivation}
In the advent of repeatedly emerging success stories in the field of artificial intelligence, for instance in medical image analysis~\citep{ker2017deep}, particle physics~\citep{bourilkov2019machine}, drug discovery~\citep{chen2018rise} or cybersecurity~\citep{xin2018machine} to mention a select few, ML algorithms have become ever more complex and sophisticated. Data-driven AI is getting more efficient and is being powered by increasingly more powerful hardware, allowing much progress to be made often surpassing human level visual cognition (see the recently coined foundation models \citep{bommasani2021opportunities} and some of their exemplary candidates \citep{ramesh2021zero}). \par
However, even though the promises of AI, especially deep learning, are big and have been expected to completely displace the work of actual humans in some fields like for example radiology \citep{obermeyer2016predicting}, reality shows that this is not the case \citep{pesapane2020myths}. The reasons for that are partly of regulatory nature but more so due to the lack of reliable reasoning of such deep learning algorithms. In the medical field where stakes are high and mistakes can result in actual damage to a persons's health, the question of \emph{intelligence} and reliable reasoning for such algorithms becomes a topic of heated discussions in the ML community. 
Only answering the questions of correlation with ML doesn't seem to be sufficient anymore but rather the question about cause and effect and understanding the decisions made by machines play a bigger role for recent advances in ML \citep{xu2019explainable, castro2020causality, scholkopf2022causality}.  \par

This lack of \emph{reasoning} becomes especially apparent in visual cognition tasks whose solution requires abstraction and composition of concepts  all while only providing a handful of samples, \emph{Bongard Problems} (BPs) \citep{bongard1970pattern} being a prime example of such tasks. Russian computer scientist Mikhail Bongard who proposed the original set of 100 BPs devised a task that is more relevant today than ever in assessing the capability of  machine computation in comparison to inferences enabled by the human brain for which causality \citep{pearl2009causality} is a crucial component, providing a way of reasoning about for BPs. To this day, they remain largely unsolved by AI/ML algorithms and even humans tend to often struggle within the extended set of BPs  \citep{mitchell2019artificial}.

\begin{figure}[ht!]
   \begin{minipage}{0.48\textwidth}
     \centering
     \includegraphics[scale=0.3]{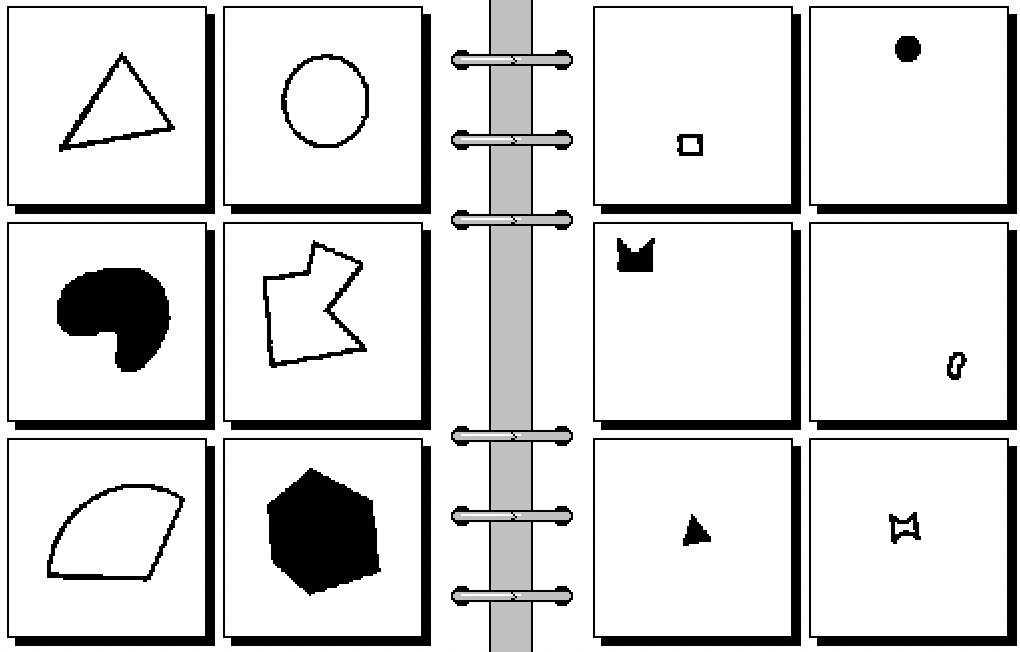}
     \caption{\textbf{BP\#2.} Solution to this BP is that in the left group all shapes are large and in the right group all shapes are small.}
     \label{Fig:p002}
   \end{minipage}
   \hfill
   \begin{minipage}{0.48\textwidth}
     \centering
     \includegraphics[scale=0.3]{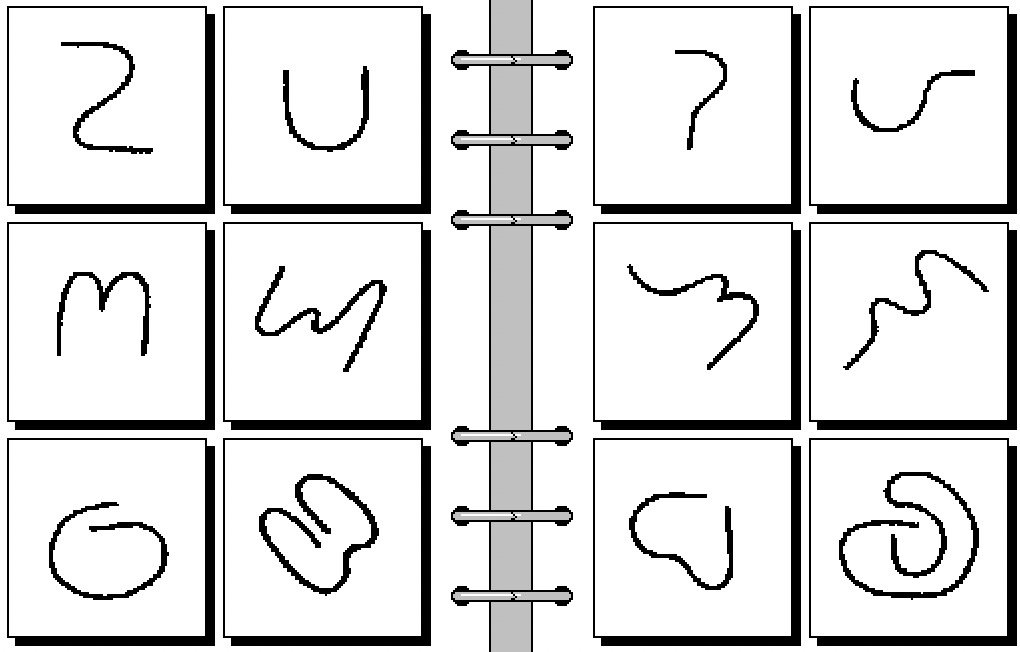}
     \caption{\textbf{BP\#72.} Solution to this BP is that in the left group the ends of the lines are parallel instead of perpendicular.}
     \label{Fig:p072}
   \end{minipage}
\end{figure}

An example for which humans could figure out the solution of a BP rather quickly compared to one for which a human would think more is in Figure \ref{Fig:p002} and Figure \ref{Fig:p072}, demonstrating the varying difficulty of BPs. \par

\subsection{Key Challenges}
BPs are a collection of images in which we are presented with two groups of images from which we have to find a set of properties present in one group but absent in the other and vice versa, e.g. Fig. \ref{Fig:p002}, where the solution is that the images on the left side are small, while the images on the right side are big. The images at first seem rather simple and elementary since they usually only consist of simple shapes and are restricted to being black and white but the underlying task requires a good sense of abstraction and pattern recognition. This can be challenging even for humans, as they reach a solution while still being incapable of recognizing how they reached said solution. Due to their compactness and very general nature, \textbf{BPs present many of today's most important ML research questions within a single framework and their solution has been described to be very close to central aspects underpinning human cognition \citep{hofstadter2006godel}}. BPs in particular seem so interesting because they provide a very clear contrast to the current state of the art for deep learning, on the one hand we have the already mentioned very sophisticated applications of deep learning on high dimensional and complex data \citep{chen2018rise, ker2017deep, bourilkov2019machine} and on the other hand we have a number of small black and white images of simple shapes and structures of which often even a child can reason from, but machines yet fail. One contributing factor to that is that many of today's visual recognition models rely on massive amounts of data \citep{he2016deep, ramesh2021zero, ker2017deep, chen2018rise}, whereas with Bongard Problems the amount of data that is available to us is very limited to only 12 images for each problem and a total of 100 problems for the original set of BPs. This being only one of the challenges we are faced with in trying to solve BPs, it is also about being able to learn symbolic concepts \citep{garcez2022neural} with very little data \citep{ravi2016optimization, snell2017prototypical} while also inferring some form of compositional reasoning \citep{battaglia2018relational} on learned concepts. \par

%% file: master_thesis/2_Related_Work.tex
\section{Related Work}
In the past there have been several attempts to solving BPs with varying success. Due to BPs encompassing multiple difficult challenges for machine learning, the approaches differ in how they try to solve them by either splitting the problem up into different sub-tasks, mainly being feature extraction then followed by pattern recognition and abstract reasoning \citep{foundalis2006phaeaco, hofstadter2006godel} or trying to solve them in an end-to-end manner where feature extraction is streamlined into the learning process \citep{nie2020bongard, kharagorgiev2020solving}. In addition to that, we also have to consider whether only the BPs themselves are used as training data, which wasn't the case in any of the approaches, or synthetic data is used for learning feature extraction \citep{nie2020bongard, kharagorgiev2020solving}. \par
In a promising effort, Depweg et al. \citep{depeweg2018solving} first translated features extracted from images into a symbolic visual vocabulary to then try to solve the BPs through a powerful formal language together with Bayesian Inference, allowing them to even output natural language as the solution, given the hand-crafted nature of the features used for this feature extraction. But consequently being limited to only solving BPs that are able to be featurized according to these hand-crafted features. After filtering out BPs which can't be expressed using this visual language, only 39 BPs from the original 100 remain of which 35 BPs were able to be solved.\par
Similar to that, Foundalis \citep{foundalis2006phaeaco} follows an evolutionary process for feature extraction with a family of different feature extractors where it can go from pixel representation to more abstract representations followed by a comparison of these features and visual pattern matching to find a satisfying solution. Succeeding in solving about a dozen BPs from the original 100. \par
Another approach is from Kharagorgiev \citep{kharagorgiev2020solving} which involves synthetically generated data of random geometric images to learn a feature extractor. The extracted features are then used to train a neural network for image classification, assigning them to the correct group for a given BP. This model was evaluated on 232 BPs including not only the 100 original BPs from Bongard \citep{bongard1970pattern} but also including the extended set of BPs \citep{mitchell2019artificial}. \par
In contrast to these approaches there has also been work on solving BPs end-to-end using deep neural networks taking images as input, this time with implicit feature extraction through neural networks \citep{nie2020bongard} but focusing more on providing the neural network with enough data to learn these features by creating synthetic data inspired by the BPs consisting of a set of 12,000 additional synthetic BPs which were then used to train a neural network, achieving an accuracy of around 60-70\% on the synthetic dataset. \par

%% file: master_thesis/3_Methodology.tex
\section{A General Perspective on BPs}
\label{sec:bps}

\begin{figure}[!t]
\centering
\includegraphics[scale=0.4]{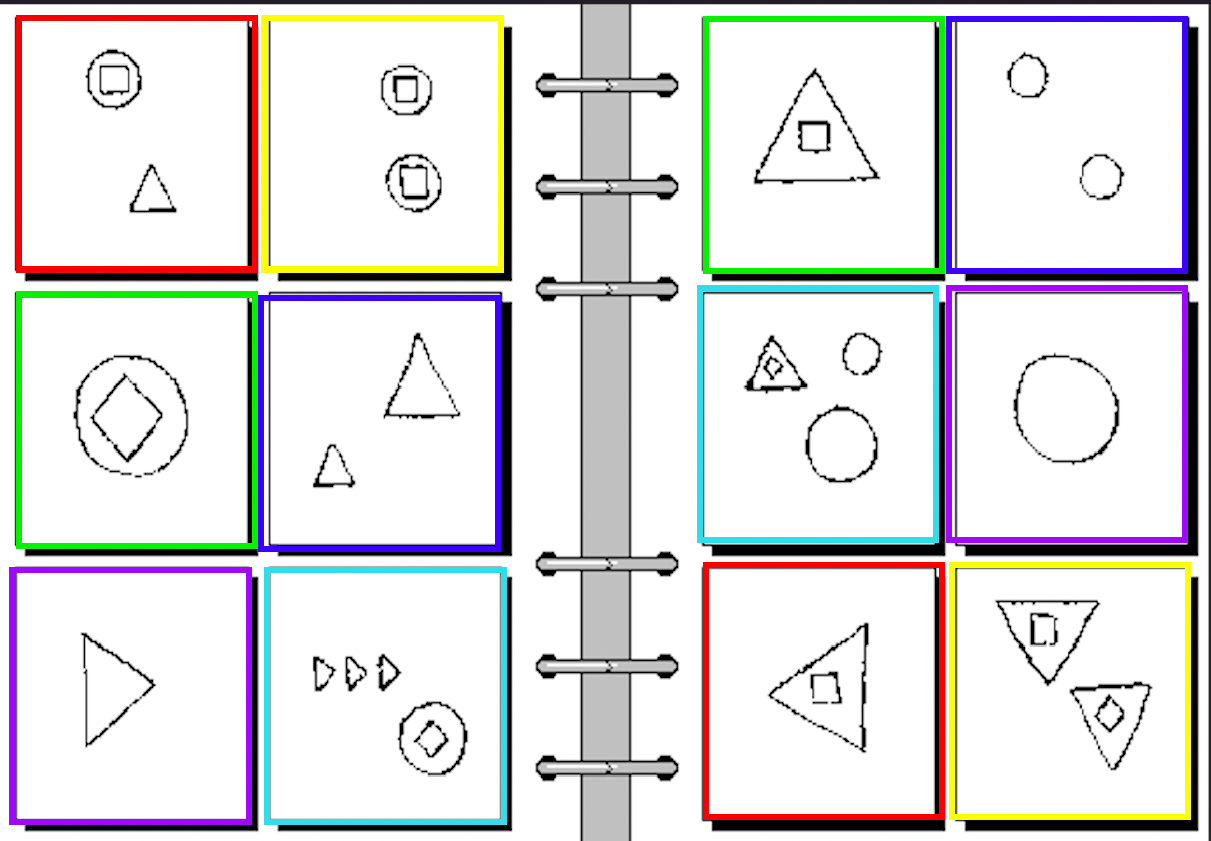}
\caption{\textbf{BP example.} An example for a BP in which we have for each image in $G_1$ a similar image in $G_2$ highlighted with a color. This shows the human bias of BP design about trying to convey a message to the solver. Each colored pair corresponds to two similar images trying to hint at or violate possible parts of the solution. (Best viewed in color.)}
\label{fig:x leading bp}
\end{figure}

A BP is a puzzle in which we are presented with a set $G$ containing 12 black and white images, this set is split into two subsets $G_1$ and $G_2$ where $G \in \mathcal{B}$, $G_1, G_2 \subset G$ and $G_1 \cap G_2 = \emptyset$. $\mathcal{B}$ is the set of all BPs with $|\mathcal{B}| = 100$. Each set contains six images, each represented as a matrix $\textbf{X}_j$ with $j \in \{1,2,...,12\}$ and $\textbf{X} \in \{0, 1\}^{w \times h}$, where $w$ and $h$ are the width and the height of the image. The goal is to find a set of separating properties $S$ which are present in all images of one group but in none of the images of the other group.  \par

The original set of BPs, designed by Mikhail Bongard \citep{bongard1970pattern},  consists of 100 puzzles. While scientists have readily increased the number of available BP puzzles to almost 400 now \citep{foundalis2006phaeaco, hofstadter2006godel}, we restrict ourselves to the original set of 100. Since BPs encode a \emph{human intention} within their respective solution, said restriction is arguably reasonable. This becomes more clear as we realize that the solution to a BP is not trivial and can be ambiguous depending on how complex we formulate the properties of $S$, multiple different solutions/interpretations for the same BP \emph{can} exist. \par
Only looking at a single image from each group is also unlikely to yield the correct solution, since BPs are designed in such a way that the correct set of properties $S$ can only be found by carefully examining common properties \emph{within a group} and contrasting properties from the other. The images are also \emph{not} generated randomly to comply with $S$, but they are often generated in a way leading the solver to the desired solution by showing violation of certain properties through very similar images, so that it is most noticeable for the human solver. We can see this in Fig. \ref{fig:x leading bp} for instance where each image with a colored border in one group has a counter part with a same colored border in the other group. These counterparts are often very similar to each other so that at least some of the desired solution properties become clear. \par
To give an exemplary walk-through for a BP, we will examine the BP in Fig. \ref{fig:x leading bp}. Do note that the borders for the images are usually not colored and are just used as a visual aid for this example.
The purple bordered images hint at a solution where different shapes \emph{could} matter, while the blue bordered images try to show that it may not be related to the numerosity of shapes in one image. The green bordered images hint at the \emph{possible} relevance of the enclosing property, again with the change of shape (change of shape seems to go from triangle to square or from square to triangle) further reinforced by the yellow bordered images which also strengthen the assumption that numerosity is indeed not important. \par
One could now think that, like numerosity, different shapes are not the deciding difference, since both sides seem to include the same set of different shapes and also the same number of shapes in one image. However, by going back to the green, yellow and red pairs, we notice that the change of shape is coupled to the enclosing property i.e., if one shape is enclosing another, on the left side it is always a circle enclosing a square and on the right side it is always a triangle enclosing a square, leaving us with the solution to this BP. This walk-through shows us that it is almost impossible to get the solution by looking at the images one by one, instead they have to be seen in combination and relation to each other, slowly revealing the solution as more images are seen. This \emph{leading} property is often observed as a form of human bias since BPs are hand-crafted by humans thus emphasizing the communication aspect of trying to convey a message to the solver.

\section{From BP to RL Environment}
\label{sec:bptorl}
Before going into the details of the RL environment setup we want to give a quick overview on why we want to frame the problem of solving BPs into an RL setting in the first place, because the decision of applying RL on a task for which the data we need is already available beforehand is rather not intuitive.
\subsection{Sequential Decision Process}
 A human trying to solve a BP usually doesn't look at the BP image as a whole but rather examines the single images from both groups separately. Also, when thinking about possible solutions for a BP a human will try to deduce sets of properties by looking at image pairs in particular because this will reveal either distinct or common properties between images, aiding in the overall process of finding the correct solution. This behaviour is shown in more detail in Section \ref{sec:bps} with the BP walk-through for the BP seen in Fig. \ref{fig:x leading bp}. \par 
With this RL environment we try to mimic this process of a human solving BPs by also looking at image pairs and comparing them. Even though it may be possible to find a solution by directly examining the BP as a whole, we aim to do so in a way that is more understandable for a human by being able to visually show the \emph{deduction} process of the learned model. Furthermore, with giving the agent the ability to take influence on the process of sampling the images and cycling through new examples, we directly exploit one of the strengths of RL in achieving better sample complexity as we reduce the search space in which we look for good examples to which we compare the images.
%\subsection{Causal Motivation}
%Another key motivation behind an RL environment is that by having actions we can bridge the gap between BPs and causality \citepp{pearl2009causality}, because now we can draw the parallel between actions and interventions, meaning that we intervene on causal factors in the BP-RL environment.

\begin{figure*}
\centering
\includegraphics[width=.8\textwidth]{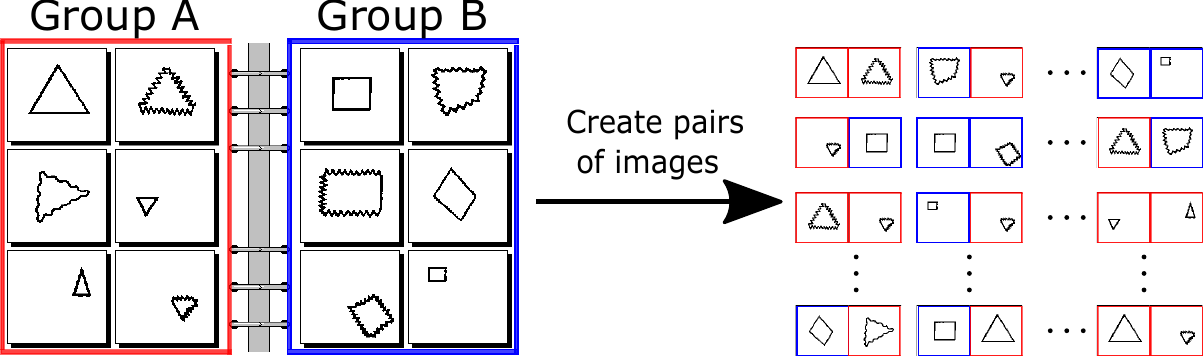}
\caption{\textbf{Data Generation.} Each image from group A (red) and group B (blue) gets split up into single images and is then combined into image pairs between both groups and within each of the groups. (Best viewed in color.)}
\label{fig:x bp to pairs}
\end{figure*}

\subsection{Design Choices for a Bongard Problem RL Environment}

To solve BPs with reinforcement learning we need to frame BPs as some kind of environment with which an agent can interact. To achieve this, we first split each BP into tuples of images $q:=(\textbf{X}_i, \textbf{X}_j)$, by taking all possible combinations of the images between the two sets and also all combinations within the same set $q \in  G \times G$. This gives a total number of $|G|^2 = 144$ pairs of two images for each BP, totaling to $|G|^2 \cdot 100 = 14400$ samples of pairs for all BPs as schematically shown in Fig. \ref{fig:x bp to pairs}. We sample from all the image combinations from a BP and decide for each pair whether the two images belong to the same group or not. \par
\begin{figure*}[!t]
\centering
\includegraphics[width=0.9\textwidth]{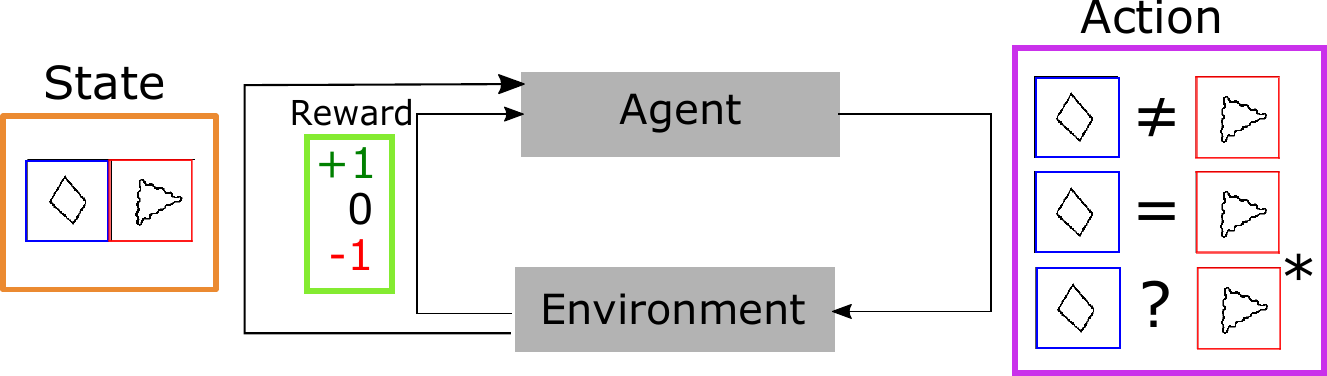}
\caption{\textbf{Decision Process.} Our RL formulation for BPs. The environment outputs a state, which corresponds to a randomly sampled image pair, the agent then can decide on whether both images of the pair belong to the same group or not, for which the agent then receives a reward, depending on the correct solution, and a new state. \textbf{*} The skip action is optional and will result in only one of the images being swapped out. (Best viewed in color.)}
\label{fig:x RL Setup}
\end{figure*}

The general RL setup for BPs is shown in Fig. \ref{fig:x RL Setup}. Let $s_t$, $r_t$ and $a_t$ be the state, reward and action at timestep $t = 1,2, .., T$ where $T$ is the number of steps in an episode. The actions correspond to assigning both images to the same group or assigning them to different groups $a_t \in \{0,1\}$ and a reward of 1 is given when the group assignment was correct and 0 if incorrect, $r_t \in \{0,1\}$. The state $s_t$ is a 2D image representation of both images where each channel corresponds to one image $2 \times w \times h$ where $w$ and $h$ are the width and the height of the image. Our goal is to find a policy $\pi(a,s)$ that maximizes the expected return $\mathbb{E}[R|s]$ where $R=\sum_{t=0}^{\infty}\gamma^t r_t$.

\begin{wrapfigure}[18]{r}{0.5\textwidth}
\centering
    \includegraphics[scale=0.6]{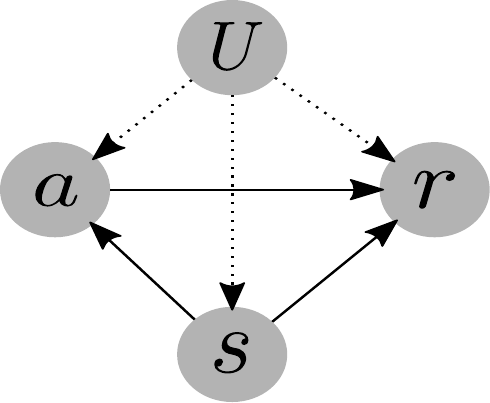}
  \caption{\textbf{Causal Graph for BPs.} We view our BP-RL setup as a Contextual Multi Armed Bandit problem. With $s$, $a$ and $r$ being the state action and reward and $U$ the unobserved confounder. The arrows represent a causal relationship and the dotted arrow refers to the unobserved confounding.}
  \label{fig:x BP SCM}
\end{wrapfigure}

\subsection{Skip Extension: The Ability to not Decide}
\label{sec:skipext}
Another extension to this environment is the addition of one more action that allows us to \emph{skip} an image. If we look at an image pair $(\textbf{X}_i, \textbf{X}_j)$ the agent now has the chance to not decide whether $\textbf{X}_i$ and $\textbf{X}_j$ belong to the same group but instead switch $\textbf{X}_j$ out for another image $\textbf{X}_k$ with $j \neq k$. Doing this gives the agent the chance to switch to image pairs for which it has more \emph{confidence}. Depending on the learned feature space the agent may be able to select relevant image pairs revealing the most information about a solution as shown in the colored borders in Fig.\ref{fig:x leading bp}. Adding the skip action will change the rewards to $r_t \in \{-1, 0, 1\}$, where the correct group assignment still gives a reward of 1, skipping gives a reward of 0 and wrong group assignment will result in a reward of -1. \emph{This skip action corresponds to the ability of the agent to form counterfactuals.} \\ In Fig.\ref{fig:x BP SCM}, we can see the causal graph of the structural causal model (SCM) $M$ from which we know that $r$ depends on $U$, $a$ and $s$. The key to understanding counterfactuals in this context lies in the statement of making a ``minimal'' modification to the current model $M$ by changing $\textbf{X}_j$, being a component of $s$, while keeping everything else the same, resulting in modified model $M_s$, referring to the ``minimal'' intervention on $s$, for which we can get the formal definition of the counterfactual: $r_s(u) := r_{M_s}(u)$. In words: the counterfactual $r_s(u)$ referring to the received reward for situation $u$ in model $M$ is defined as the reward received in the modified model $M_s$.

\subsection{Causal Assumptions for BPs}
\label{sec:causalbp}

Besides formulating the problem in an RL setting we also need to formalize the problem in a causal setting since we want to pose causal assumptions on the data generating process. We can thus approach the problem as a \emph{Contextual Multi Armed Bandit} (CMAB) because the actions we take do not influence the state, if we leave out the skip extension, this easily translates into a Structural Causal Model (SCM) \citep{pearl2009causality} which describes the causal relations between the variables of interest through functional relationships. Our causal assumptions are depicted in Fig. \ref{fig:x BP SCM}.\par
We describe the idea behind a BP as the intention $U$ which is an unobserved confounder since we don't now the intention behind a BP during learning. We have also already described the task of solving BPs as a communication problem, and here we see how the message Bongard is trying to convey i.e. intention, directly influences the state of image pairs $s$, the reward $r$ we give out and the action $a$ we take during each time step. The reward is based on the action we take and whether it violates the intention of the BP. Lastly, the intention is the only confounder and source of the data generating process behind all the examples in a BP.

%% file: master_thesis/5_Analysis_Part_II.tex
\section{Empirical Analysis: What did the agent learn? }
After analyzing many aspects of the setup of the framework in which we want to solve BPs and showing that there is \emph{something} that is learned, we now want to discuss in this chapter \textbf{what exactly is learned}. Answering this question is no easy task, even more so because the task we are trying to solve is only a proxy task to the original BP task we want to solve. In the original BP task, one is supposed to define a set of properties that is present in one group of a BP but not in the other, this set of properties is not predefined and up to knowledge of the solver about shapes, relations and meaning of geometrical shapes. Our agent, however, doesn't posses such external knowledge and has no outside grounding of the concepts it learns. The agent only makes binary decisions on whether a pair of images belong to the same group or different groups. Evaluating these decisions and analyzing whether the learned decision boundary is according to the BP solution as intended by Bongard requires some form of grounding or interpretation by a human. This of course is vulnerable to human bias being introduced into the analysis of these results but in this case is unavoidable as there is no predefined set of properties to choose from. We will start by looking at a ranking of BPs on which the agent performed well, then followed by an analysis of the learned feature space. After looking at different interpretation for the feature space we will look on some exemplary decision sequences of the agent and their interpretation.

\begin{tcolorbox}[width=\linewidth, halign=left, colframe=black, colback=white, boxsep=0mm, arc=2mm] 
\textbf{Investigating the reasons of what set of properties separates the BPs the agent performed well on from the ones with bad performance by looking at these images, almost seems like a new kind of meta BP. Highlighting the difficulty of BPs, not being about recognition or classification but more about compositional reasoning and abstraction.}
\end{tcolorbox}

\section{BP Rankings}
\begin{figure}[ht!]
\centering
\includegraphics[scale=0.95]{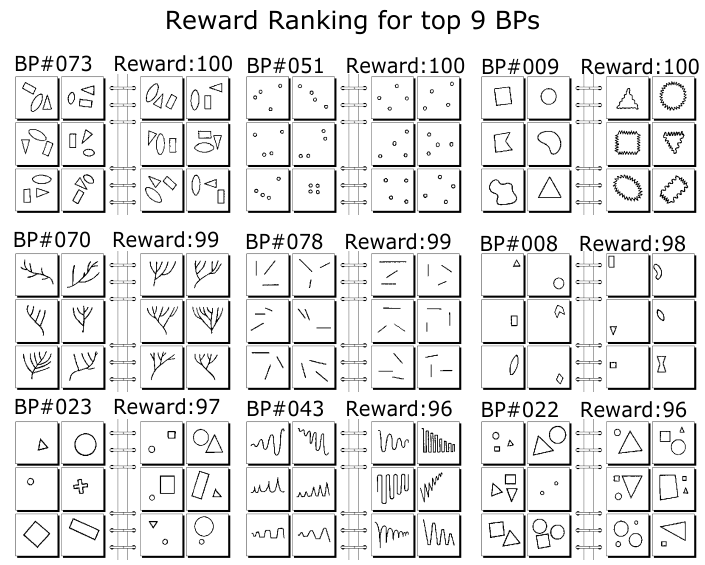}
\caption{\textbf{BP Ranking Top 9.} A ranking of the top 9 BPs according to achieved reward per BP. Reward of 1 for correct classification and -1 for wrong classification for a sequence of 100 pairs. Skipping gives a reward of 0.}
\label{fig:bp ranking}
\end{figure}

In the experiments the agent was able to learn good feature representations for about 27 of the 100 BP of which in Figure \ref{fig:bp ranking} we can see the 9 BPs on which the agent performed best on. At first glance there doesn't seem to be a common property which all of these BPs share that would make an easy identification trait for BPs with good performance. To get a better grasp on what these BPs may have in common it is also necessary to look at the solutions for each BP. Additionally, we will also plot the feature representation of the good performing BPs by applying dimensionality reduction with t-SNE \citep{van2008visualizing}.

\section{Feature Space of BPs}
\begin{figure}[h!]
\centering
\includegraphics[scale=0.75]{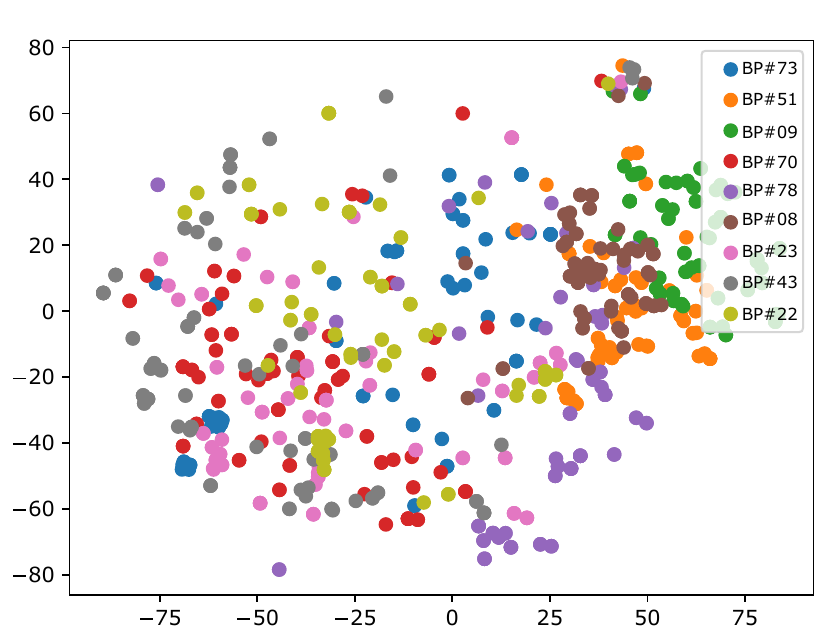}
\caption{\textbf{BP Feature Space.} A 2-d plot of the t-SNE reduced feature space for the top 9 BPs the agent performed best on. Each point corresponds to one image pair, where the feature vector is the absolute difference between each individual image encodings. The color of each point indicating to which BP an image pair belongs to. (Best viewed in color.)}
\label{fig:bp feature space}
\end{figure}

Now by combining the information about well performing BPs, their solution and the corresponding feature space we can slowly deduce the meaning of learned feature space.
Figure \ref{fig:bp feature space} shows us a two dimensional plot in which each point corresponds to a comparison of two images from a single BP in which we can observe a few clusters in which points for a single BP are very close to each other (brown, orange, green), also many of the BPs are more spread out and don't form clusters within a BP. It is very important to keep in mind that the goal is not to classify according to BPs but rather classify according to concepts and features, meaning that dense clusters for BPs are not necessarily the goal. \par

\subsection{Numerosity as Solution Space}
\begin{figure}
\centering     %%% not \center
\subfigure[\textbf{BP\#23.} Solution to this BP is that in the left group there is only one figure and in the right group there are two figures.]{
\label{Fig:p023}
\includegraphics[width=60mm]{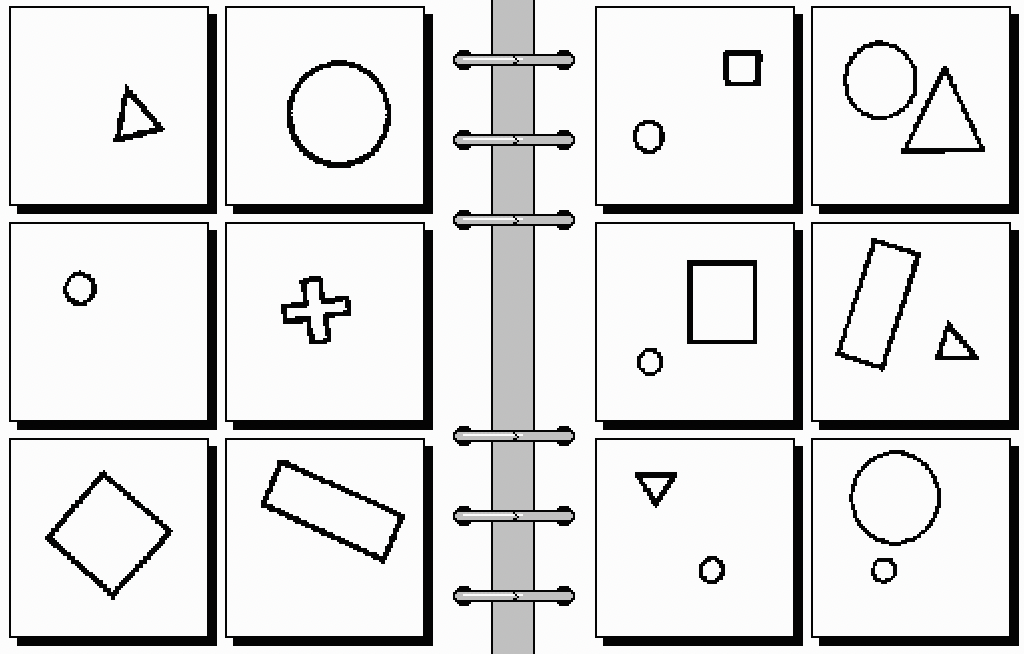}}
\subfigure[\textbf{BP\#70.} Solution to this BP is that in the left group there are no side branches of the second order and in the right group there are branches of the second order.]{
\label{Fig:p070}
\includegraphics[width=60mm]{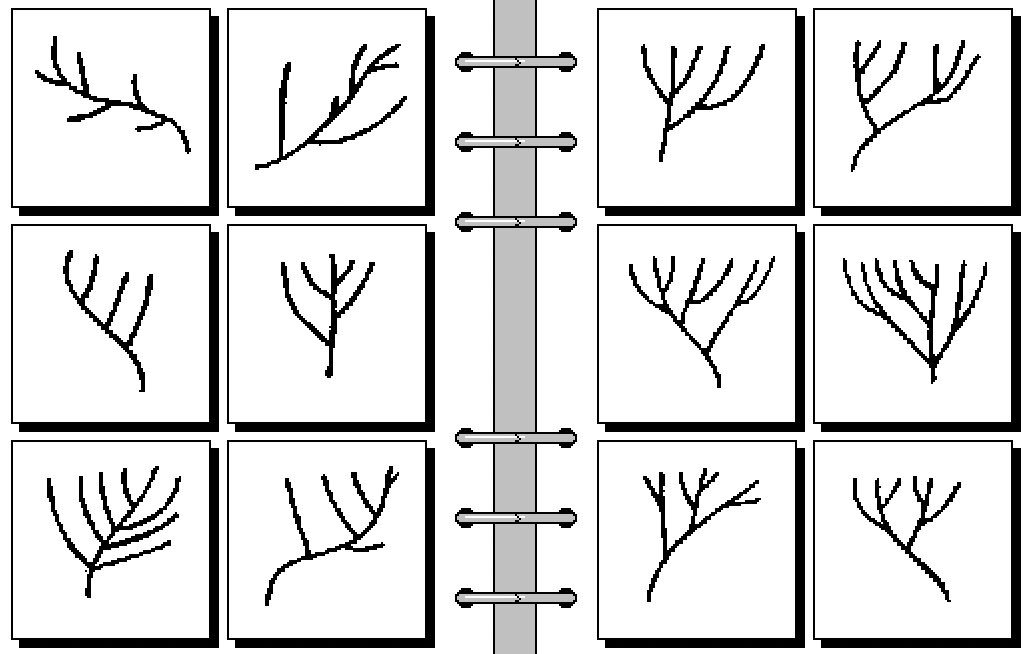}}
\end{figure}

What we can examine, giving us more insight into these feature space, is which points or which BPs in particular are close to each other. One example is BP\#23 (pink) in Figure \ref{Fig:p023} and BP\#70 (red) in Figure \ref{Fig:p070}, for which we can see that the pink and red dots, even though they are spread out, are close to each other when they appear. Both these BPs look very different, one consists of shapes like circles and rectangles and the other consists of branching lines, resembling a twig, so what do they have in common? By comparing the solutions for both BPs, we can note that BP\#23 only has one figure on the left side and two figures on the right side. BP\#70 hast only one main branch on the left side and two or more on the right side. So the similarity for both BPs seems to be in the space of solutions for which the agent seems to have learned some concept for numerosity, helping to solve these two BPs very well and \textbf{showing even though the images don't resemble each other, their resemblance lies in the shared solution space.} More examples for the interpretation of the feature space, including change of size and positional encoding, can be found in the appendix. \par

\section{Decision Sequences}
With the introduction of the skip-action the agent has the ability to not decide on every image pair whether they belong to the same group or not but instead skip the pair and pertubate one of the images to exchange it for another image. This allows the agent to decide on pairs it is most confident in and also possibly make the decision for a human more explainable by observing the sequence of decisions the agent made.
In this section we will take a look at a few decision sequences for BPs in which the agent performed well in but not perfectly, since in the cases where the agent performs too well not many skips occur. More example decision sequences can be found in the appendix.

\subsection{Example Sequences BP\#23}
\begin{figure}
\centering     %%% not \center
\subfigure[\textbf{BP\#23 First Skip.} The agent comparing a small circle and a cross, choosing the skip acting to swap out the cross.]{
\label{Fig:p023 1}
\includegraphics[width=60mm]{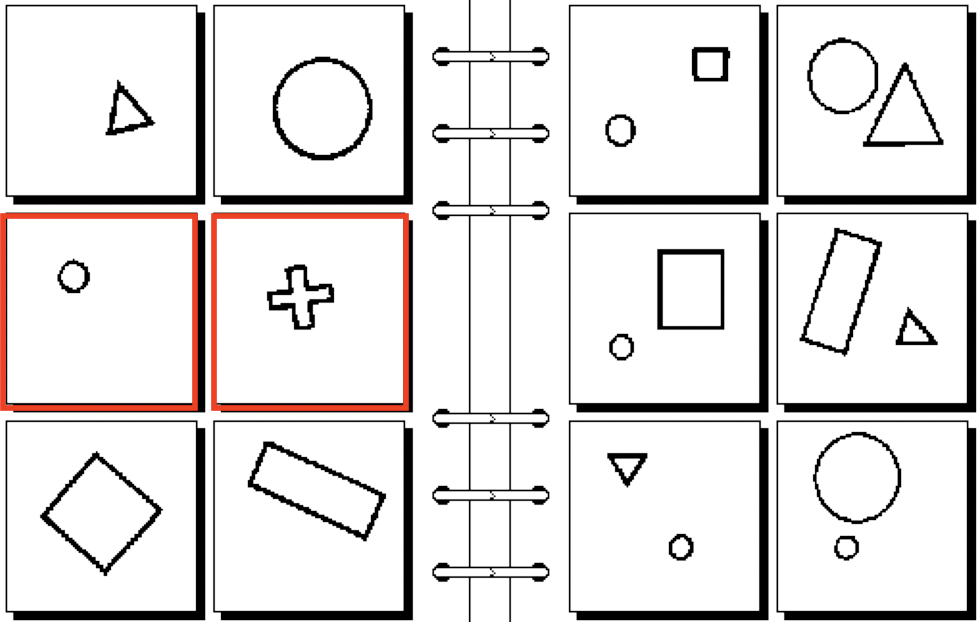}}
\vspace{.5cm}
\hspace{.5cm}
\subfigure[\textbf{BP\#23 Second Skip.} The agent comparing a small circle and a rectangle, choosing the skip acting to swap out the rectangle.]{
\label{Fig:p023 2}
\includegraphics[width=60mm]{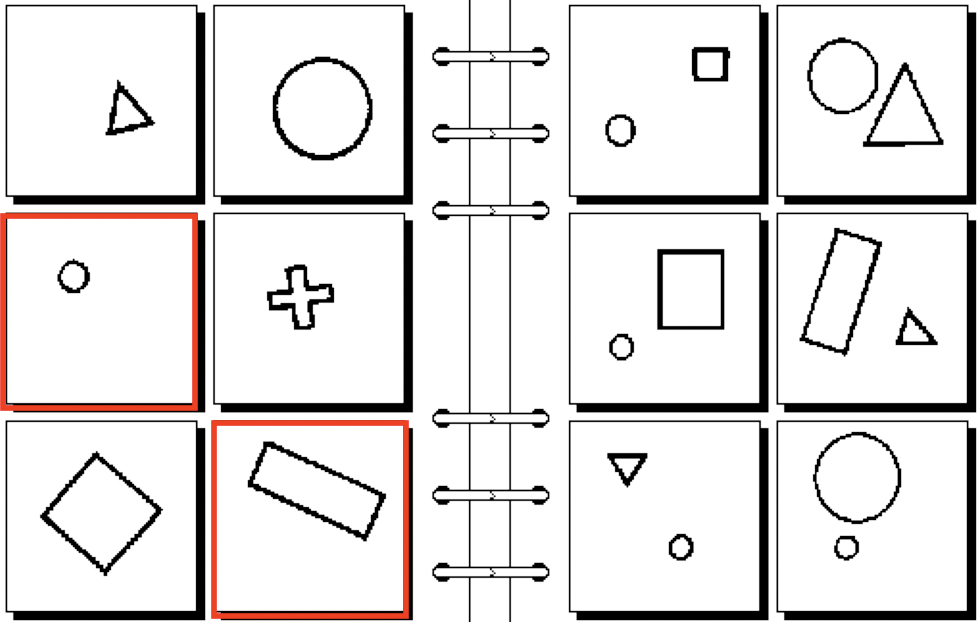}}
\vspace{.5cm}
\subfigure[\textbf{BP\#23 Successful Guess.} The agent comparing a small circle and a small circle together with a big circle, guessing correctly.]{
\label{Fig:p023 3}
\includegraphics[width=60mm]{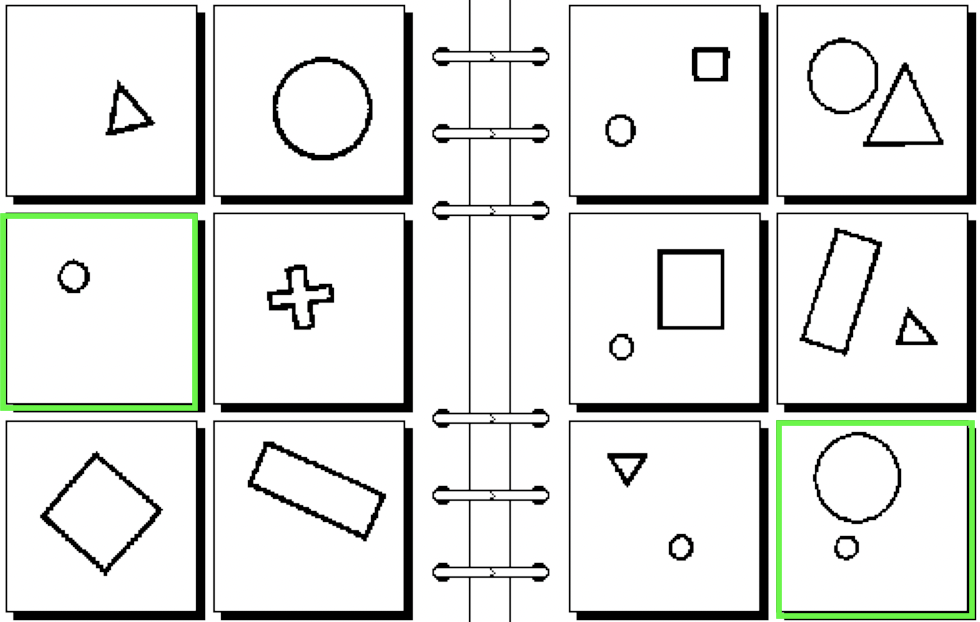}}
\hspace{5cm}
%\end{figure}
%\begin{figure}
\centering
\subfigure[\textbf{BP\#43 First Skip.} The agent comparing a more rounded wave and a spiky wave, where the rounded wave gets switched out for another spiky wave in the opposing group.]{
\label{Fig:p043 1}
\includegraphics[width=60mm]{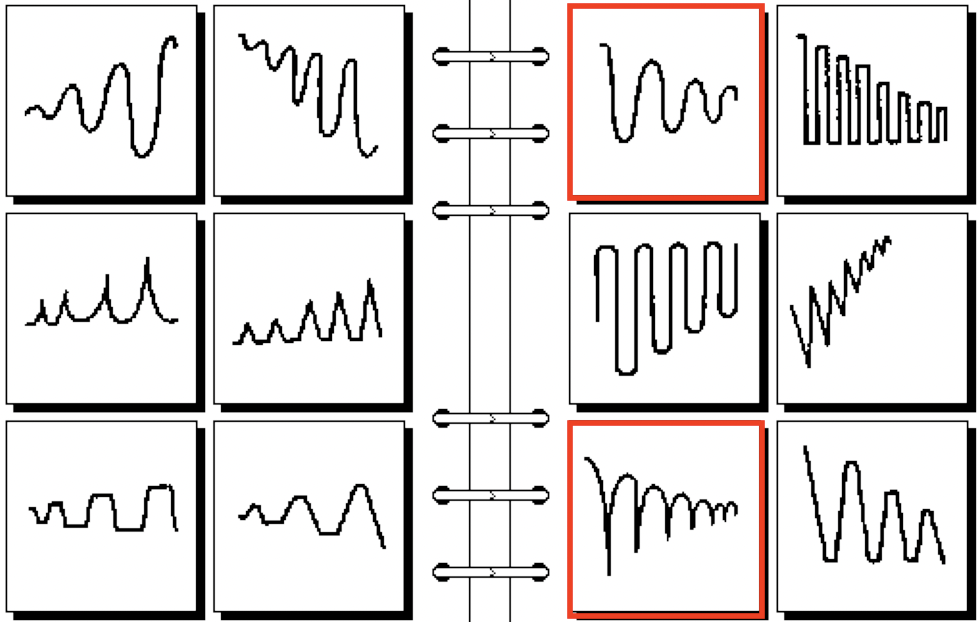}}
\hspace{.5cm}
\subfigure[\textbf{BP\#43 Successful Guess.} The agent comparing a spiky wave with another similar but inverted spiky wave, resulting in a correct guess.]{
\label{Fig:p043 2}
\includegraphics[width=60mm]{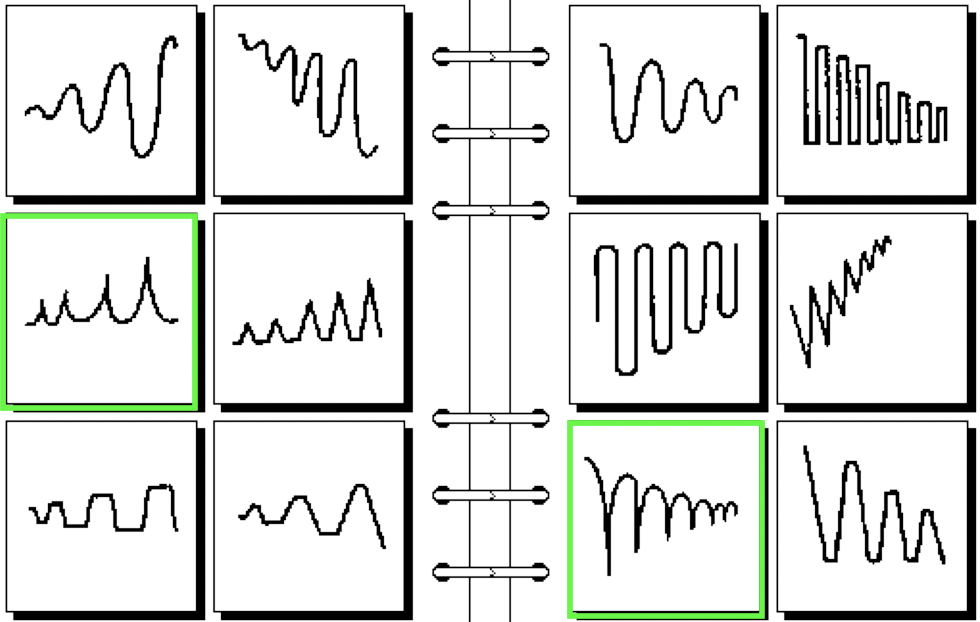}}
\end{figure}

For this BP we want to look at examples in which the agent decides to skip a pair, even though the agent performs well on this BP, for the process of learning it is very useful to compare pairs from which it can extract the most information from. Similar behaviour was demonstrated in Figure \ref{fig:x leading bp}, showing some properties of BPs giving us some examples in pairs to better lead to a solution. Figures \ref{Fig:p023 1}, \ref{Fig:p023 2}, \ref{Fig:p023 3} are the first sequence in which the agent decides to skip a comparison for BP\#23. The comparison being a small circle and a cross for which the cross gets swapped out with a rectangle, which in turn is swapped out again for the image with the small circle together with a big circle. By having a comparison most similar to the image itself but yet different enough to warrant the assignment to the other group, it is easier for the agent to learn meaningful representations for the images. \\

\subsection{Example Sequences BP\#43}
BP\#43 also has interesting decision sequences, one example being the sequence shown in Figure \ref{Fig:p043 1} and \ref{Fig:p043 2}. In this example we are looking at waves of different shapes and amplitudes, either increasing from left to right side or decreasing from left to right. The skip here is interesting because the first comparison is between a more rounded wave and a spiky wave, where the rounded wave gets switched out for another spiky wave in the opposing group. If one were to pick out the most similar wave for the spiky wave on the right side, one would pick the same wave as the agent did, because they look very similar but are opposing in the applied context. This also refers to this leading property of BPs, designing the images on the left and right side could in theory just be according to the rule of ``left to right amplitude increase'' and ``left to right amplitude decrease'' but Bongard rather decided on often picking waves from one side and inverting the applied concept, to make deducing the solution easier, on which the agent also picked up.

%% file: master_thesis/6_Conclusion.tex
\vspace{-.25cm}
\section{Discussion and Future Work}
This paper covered a variety of additions for solving problems in the realm of BPs, a field in which research in the past years was easily overlooked due to its deceivingly simple appearance. It almost seems contradictory going back to applying ML concepts to images with squares and circles while at the same time it is possible to auto generate high resolution artworks from only a few words \citep{ramesh2022hierarchical}. We first formulated BPs into an RL setting. Even though taking this step is not so obvious at first it allowed us to pose more insightful questions and also provided the tools for further investigation for meaning. 
Furthermore, driven and inspired by the concepts of causality, adjustments to the RL environment like introducing a skip action gave the agent the possibility to form counterfactuals, the question of what if the agent were to choose another image to compare to, given the same initial situation. With the skip action it is often possible to even follow the deduction process of the agent in a similar way a human would choose pairs to deduce certain properties for a given BP. In addition to the RL framing, SNNs were applied for comparing the BP generated image pairs, bringing the problem closer to the question of learning concepts by comparison. One of the most interesting results presented in this paper is the interpretation of the learned feature spaces. By analyzing the feature space through dimensionality reduction, we were able to hypothesize possible explanations for the structure of the feature space making up the right features to form a space of solutions involving these features. This success in feature representations was for approximately only a third of all BPs with many not being learned at all. But it is still promising progress for which even more BPs will be able to be solved by further improvements.
Future work may involve general improvement of the model architecture by for example including temporal convolutions on the sequence of inputs or including an attention mechanism \citep{vaswani2017attention}. More extensive analysis of the learned feature spaces and their interpretation may also provide answers in why the agent failed on certain BPs. Analyzing each BP of all the original 100 BPs, designed by Bongard may yield more information of what is easily captured by the current model architecture. Also, extending the dataset included in the BP-RL environment to more BPs, for which there are almost 400 of \citep{mitchell2019artificial}, may reveal another layer of complexity by showing how much of a communication problem BPs are, since we are switching between different creators (communicators) of BPs, to which an agent may overfit on the communication style. Analyzing this and generalization to new BPs is a very interesting direction for further research. \par 
Conclusively, this paper provided many initial steps for which further research can be pursued individually for each, with the main intention being to once again draw attention to the challenge of solving BPs in a time where problems like these get overshadowed by advances in designing highly complex model architecture and computationally intensively trained models achieving very impressive results.

%% file: master_thesis/appendix.tex
\clearpage
\section{Change of Size as Solution Space}
Another very interesting example are the BP\#43 (grey) and BP\#22 (lime), again both BPs don't look anything alike. One consists only of one line forming some kind of shape and the other consisting of multiple shapes like triangles, circles and squares. We will therefore again compare both solutions. For BP\#43 the amplitude increases from left to right in the left group and in the right group it decreases from left to right. For BP\#22 the shapes in the left group roughly have the same size while the shapes in the right group have very different sizes. In contrast to the previous example, the solution space here doesn't seem to be similar, but yet they are close to each other and also perform well. However, this example is so interesting because if we take a closer look at BP\#43 we can notice that the change in amplitude on the left side is (mostly) not so significant if compared to the change of amplitude on the right side, where the change of amplitude seems a lot bigger than on the left. BP\#22 having small changes in area (from triangle to square) on the left side and big changes in area on the right side, we have again found a similarity in the solution space.

\begin{figure}[h!]
\centering     %%% not \center
\subfigure[\textbf{BP\#43.} Solution to this BP is that in the left group the amplitude increases from left to right and in the right group it decreases from left to right.]{
\label{fig:a}
\includegraphics[width=60mm]{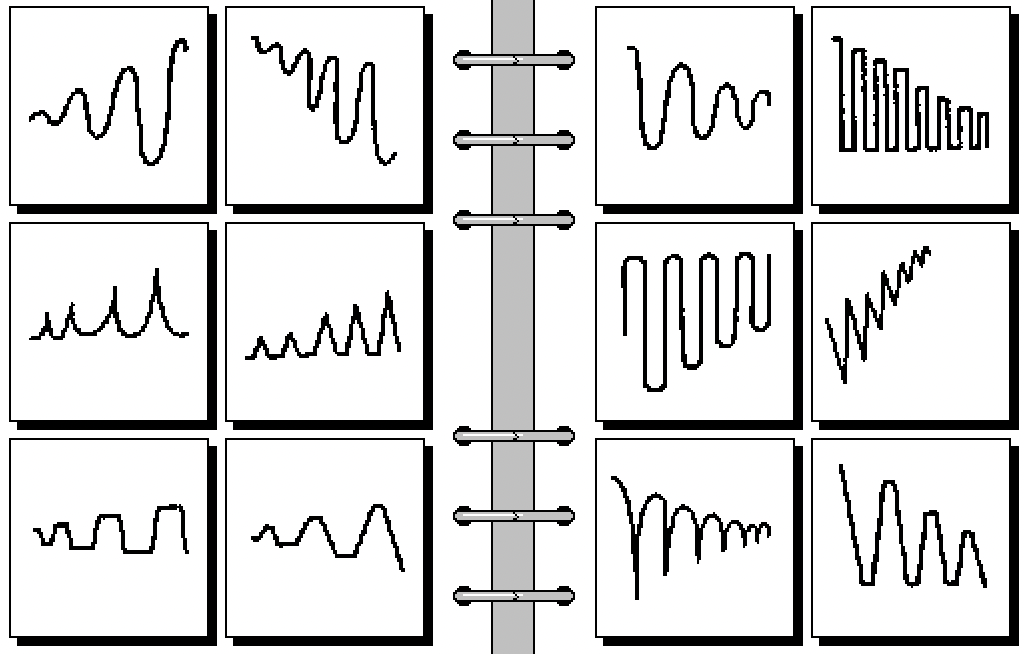}}
\subfigure[\textbf{BP\#22.} Solution to this BP is that in the left group there are shapes of roughly the same size and in the right group the shapes differ greatly in size.]{
\label{fig:b}
\includegraphics[width=60mm]{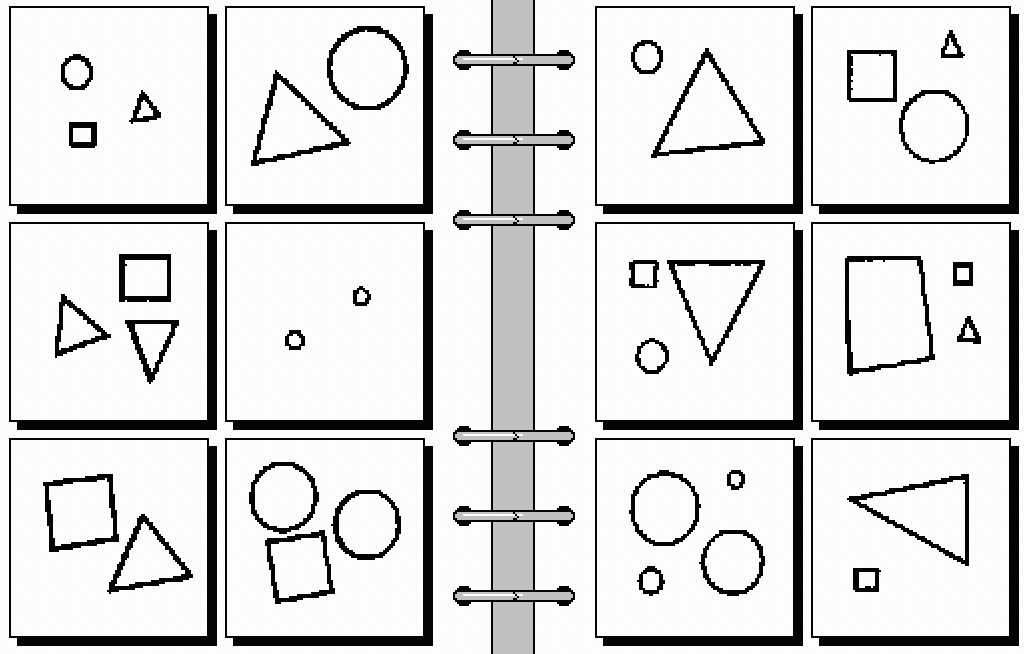}}
\end{figure}

\section{Positional Encoding in Solution Space}
In Figure \ref{fig:bp feature space} on the right side of the plot there is a noticeable cluster of brown, orange and purple dots where the overlap, with green being more outside the cluster. This hints at another common property in the solution space. As observed in the previous two examples, again the images don't look alike. One being dots, the second being lines and the third small simple shapes. So the focus is on the solution space in which we want to find similarities between these images. The solution for BP\#51 is that in the left group there are always circles close to each other and on the right side there are no circles close to each other. For BP\#78 it is that in the left group the position of the lines is so that the extensions of the lines would cross in one point and in the right group the extension of the lines would not cross in one point. For BP\#8 the solution is that in the left group all shapes are on the right side of the image and in the right group all the shapes are on the left side of the image. There again doesn't seem to be an obvious similarity between these solutions but something they all have in common, after close inspection, is that they are positional encodings. This means that getting to the solution has to involve knowledge about the position of the images. They are also the only BPs in which positional encoding is relevant in the best 9 performing BPs, further strengthening this hypothesis.

\begin{figure}[h]
\centering     %%% not \center
\subfigure[\textbf{BP\#51.} Solution to this BP is that in the left group the there are always 2 circles close to each other and in the right group no circles are close to each other.]{
\label{fig:a}
\includegraphics[width=60mm]{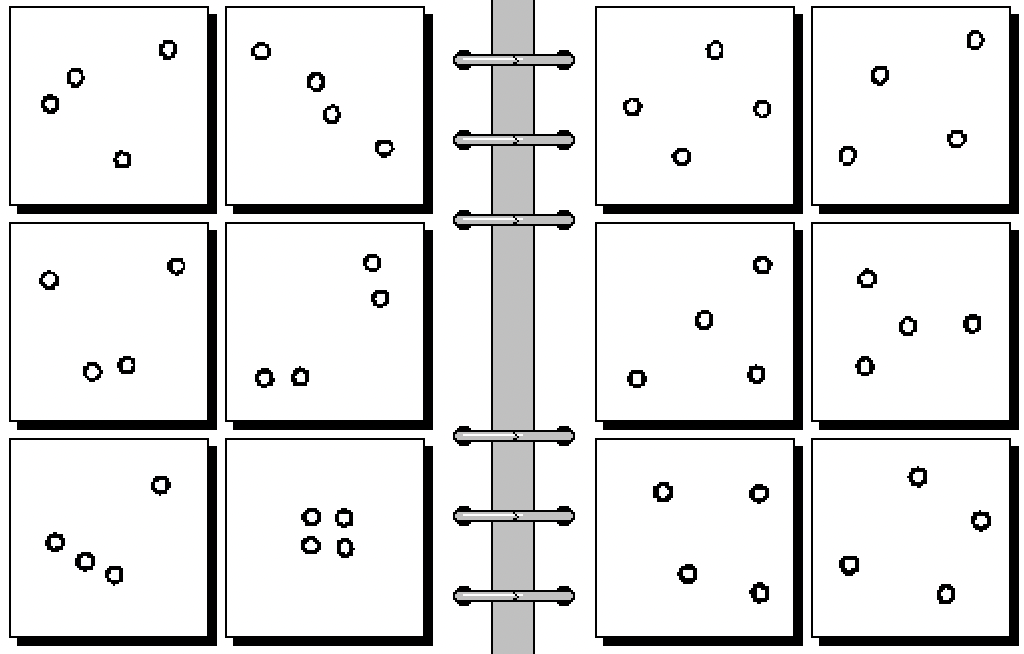}}
\subfigure[\textbf{BP\#8.} Solution to this BP is that in the left group the shapes are all on the right side and in the right group the shapes are all on the left side.]{
\label{fig:b}
\includegraphics[width=60mm]{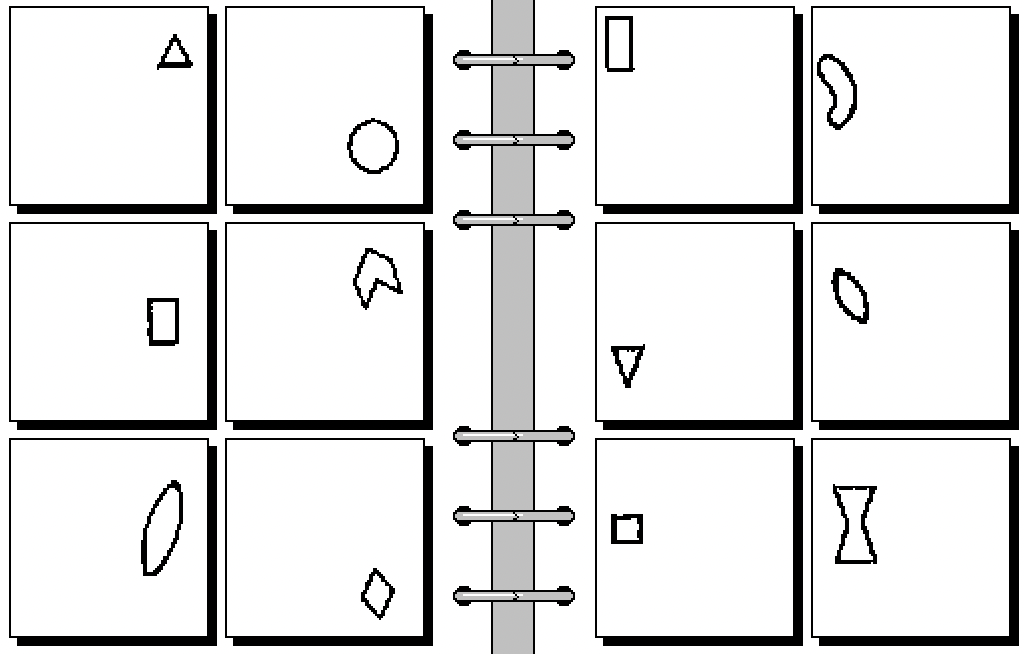}}
\subfigure[\textbf{BP\#78.} Solution to this BP is that in the left group the extension of the lines cross at one point and in the right group the extension of the lines do not cross at one point.]{
\label{fig:b}
\includegraphics[width=60mm]{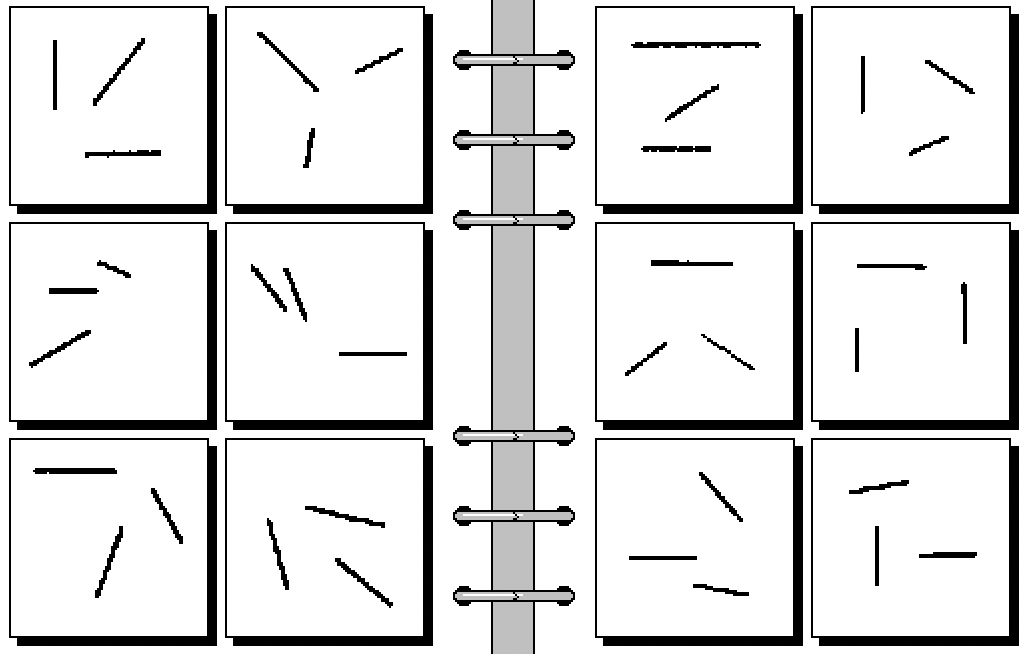}}
\end{figure}

\clearpage

\section{Example Sequences BP\#23}
\begin{figure}[h!]
\centering     %%% not \center
\subfigure[\textbf{BP\#23 Skip.} The agent comparing a small circle and a cross, choosing the skip acting to swap out the cross.]{
\label{Fig:p023 4}
\includegraphics[width=60mm]{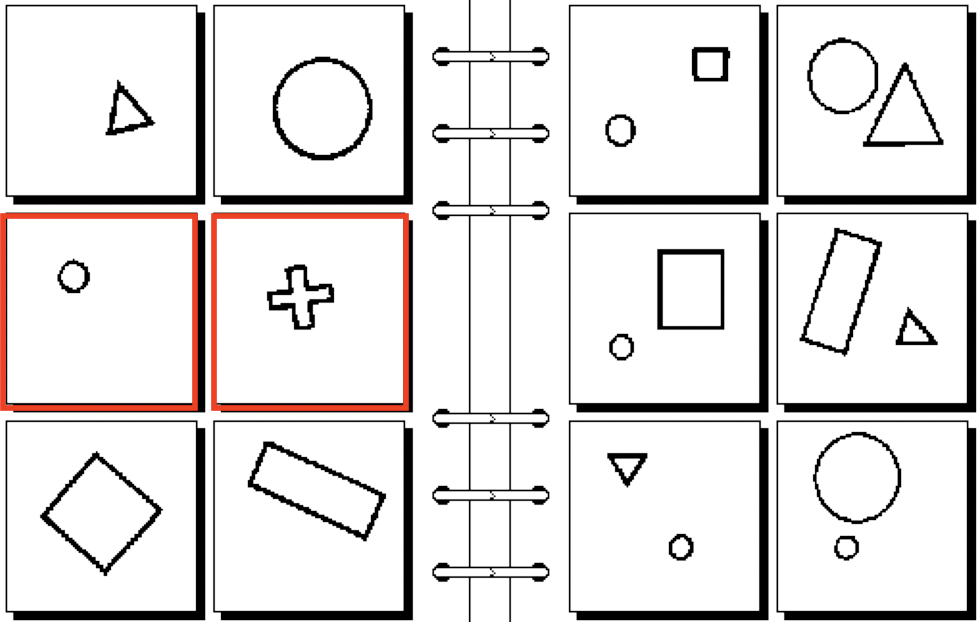}}
\subfigure[\textbf{BP\#23 Successful Guess.} The agent comparing a small circle and a another small circle together with a rectangle, guessing correctly.]{
\label{Fig:p023 5}
\includegraphics[width=60mm]{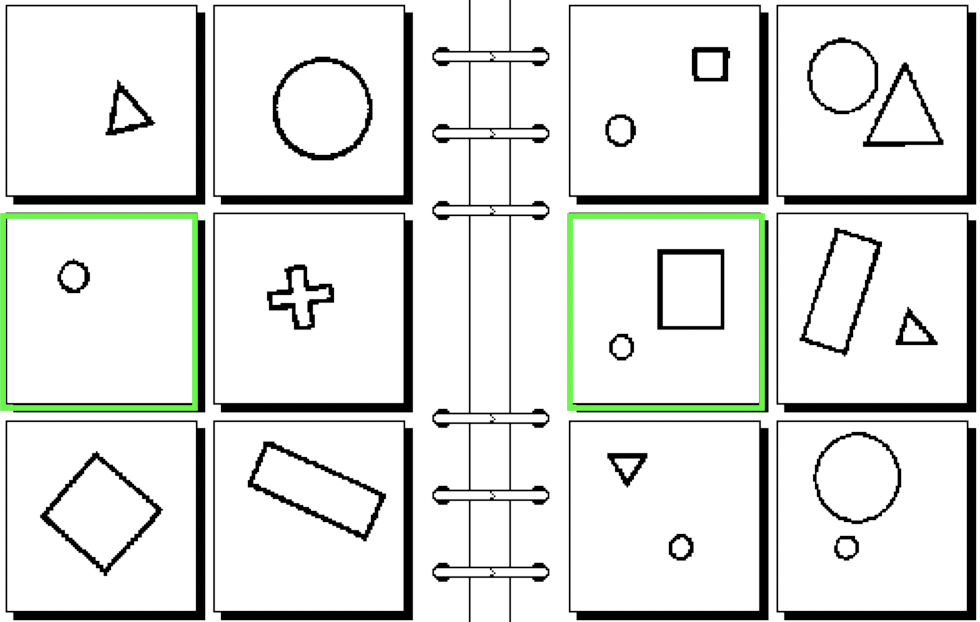}}
\end{figure}

Another example sequence for BP\#23 showing the same decision behaviour is shown in Figure \ref{Fig:p023 4} and \ref{Fig:p023 5},
again the agent decides to skip the cross comparison to switch it out for an image from
the opposing group in which the circle also appears together with a rectangle, resulting in
a correct guess.
59

